\documentclass[12pt]{article}

\newcommand{\ts}{\textsuperscript}

\usepackage{graphicx}
\usepackage{caption}
\usepackage{subcaption}
\usepackage{verse}
\usepackage{textcomp}
\usepackage{amsmath}
\usepackage{endnotes}
\usepackage{hyperref}
\hypersetup{
    colorlinks,
    citecolor=black,
    filecolor=black,
    linkcolor=black,
    urlcolor=black
}

\let\footnote=\endnote

\title{A Computational Approach to Walt Whitman's Stylistic Changes in \textit{Leaves of Grass}}
\author{Jieyan Zhu\\ Columbia University \\ \texttt{jz2988@columbia.edu}}
\date{}

\begin{document}

\maketitle
\begin{abstract}
This study analyzes Walt Whitman's stylistic changes in his phenomenal work \textit{Leaves of Grass} from a computational perspective and relates findings to standard literary criticism on Whitman. The corpus consists of all 7 editions of \textit{Leaves of Grass}, ranging from the earliest 1855 edition to the 1891-92 ``deathbed'' edition. Starting from counting word frequencies, the simplest stylometry technique, we find consistent shifts in word choice. Macro-etymological analysis reveals Whitman's increasing preference for words of specific origins, which is correlated to the increasing lexical complexity in \textit{Leaves of Grass}. Principal component analysis, an unsupervised learning algorithm, reduces the dimensionality of tf-idf vectors to 2 dimensions, providing a straightforward view of stylistic changes. Finally, sentiment analysis shows the evolution of Whitman's emotional state throughout his writing career.
\end{abstract}

\section{Introduction}
Do writers exhibit stylistic changes as they reach old age? For most writers it is a hard question. People write works in different genres, with various themes. It is therefore tedious to isolate their style and conduct research without being affected by other literary elements in their works. However, it seems not the case for Walt Whitman. Whitman, best known for his \textit{Leaves of Grass}, is an icon of American poetry. There were 7 U.S. editions of this classic published throughout Whitman's lifetime, including the 1891-1892 ``deathbed'' edition, which was merely a reprinting of the 1881-1882 edition with 2 ``Annexes'' attached to the end. Some Whitman scholars argue that the 1891-1892 \textit{Leaves of Grass} is not an edition.\footnote{For example, see Amanda Gailey, ``The Publishing History of \textit{Leaves of Grass},'' in \textit{A Companion to Walt Whitman}, ed. Donald D. Kummings (Malden: Blackwell Publishing, 2006), 409-438.} Considering that the 2 ``Annexes'' still account for about 10\% of the text, we include the 1891-1892 \textit{Leaves of Grass} as a separate edition in our analysis. 

\par
The debut of \textit{Leaves of Grass} was in 1855, when Whitman was 36 years old. Whitman spent the rest 37 years of life revising and adding poems, turning a thin pamphlet into a weighty tome of the 19\ts{th} century American history. Given the long timespan of publication history, \textit{Leaves of Grass} is excellent for a computational study of how a writer's style changes in his/her career. 

\par
Most Whitman scholars hold the view that Whitman lost his ability to produce works of poetic innovation and aesthetic worth in his late career. However, some recent studies have found that there is still the presence of a ``radical, rebellious, and Whitmanian signature style'' in the last editions of \textit{Leaves of Grass}, generating crucial contributions to the shaping of modernist innovations and achievements after Whitman's death.\footnote{Caterina Bernardini and Kenneth M. Price, ``Whitman’s `Deathbed' Radicalism and Its Modernist Effects,'' in \textit{The New Walt Whitman Studies}, ed. Matt Cohen (Cambridge: Cambridge University Press, 2019), 17-32.} Anyway, there is no doubt that Whitman underwent astounding stylistic changes during his lifetime. As the famous Whitman scholar C. Carroll Hollis once said: ``I cannot believe that a change comparable to that in Whitman's diction can ever be found.''\footnote{C. Carrol Hollis, \textit{Language and Style in Leaves of Grass} (Baton Rouge: Louisiana State University Press, 1983), 220.} In this study, a variety of computational techniques are applied with the goal to quantify Whitman's stylistic changes in ``Leaves of Grass'' computationally, which hopefully will help Whitman scholars gain new insights into this gem of American poetry.

\section{Corpus}
The Walt Whitman Archive stores almost all of Whitman's writings during his career.\footnote{Almost everything about Whitman can be found in Matt Cohen, Ed Folsom, and Kenneth M. Price, ``The Walt Whitman Archive,'' University of Nebraska-Lincoln, accessed August 25, 2021, \url{https://whitmanarchive.org/}.} All 7 U.S. editions of \textit{Leaves of Grass} are available for download in extensible markup language (XML) format. The contents are then extracted with the \verb|BeautifulSoup| function in the \verb|bs4| package of Python.\footnote{Leonard Richardson, ``Beautiful Soup 4.9.0 documentation,'' accessed August 25, 2021, \url{https://www.crummy.com/software/BeautifulSoup/bs4/doc/\#}.} The line groups of \textit{Leaves of Grass} are classified by \verb|type="linegroup"|, \verb|type="poem"|, \verb|type="section"|, etc., as did by Whitman in his original manuscripts. This variety of structures in XML files reflects the richness of poetic forms in \textit{Leaves of Grass}, and plain text files (TXT) are simply not enough for a complete description of such a variety. 

\par
The first difference among editions is that later editions are longer than earlier ones. As shown in Table \ref{tab::stanzas}, the number of stanzas increases, except for the 1871-72 edition, where the number of stanzas drops a bit. In the end, the deathbed edition (1891-1892) is almost three times the size of the first edition if changes in the average stanza size are not considered. The 1881-82 edition is the single largest expansion of \textit{Leaves of Grass} since Whitman viewed it as the ``definitive'' edition of Leaves of Grass, the unification of his works altogether.\footnote{Dennis K. Renner, ``\textit{Leaves of Grass}, 1881–82 edition,'' in \textit{The Routledge Encyclopedia of Walt Whitman}, eds. J.R. LeMaster and Donald D. Kummings (New York: Routledge, 1998), 373-375.} More statistics on stanzas is in the next section.

\begin{table}[h!]
\centering
\begin{tabular}{c|c}
\hline
Edition & Number of Stanzas \\
\hline
1855 & 643 \\
\hline
1856 & 969 \\
\hline
1860-61 & 1,418 \\
\hline
1867 & 1,627 \\
\hline
1871-72 & 1,381 \\
\hline
1881-82 & 1,948 \\
\hline
1891-92 & 2,072 \\
\hline
\end{tabular}
\caption{Number of stanzas in \textit{Leaves of Grass}}
\label{tab::stanzas}
\end{table}

\section{Stanzas}
Whitman is known to experiment with stanza forms when compiling \textit{Leaves of Grass}. For example, Whitman numbered all stanzas in the 1860-61 edition, and he numbered all sections in the 1867 edition. Subsequently, he deleted stanza numbers but preserved section numbers in the 1881-82 edition. In addition, he introduced a new poetic structure called ``cluster'' starting from the 1860 edition.\footnote{For a discussion of poetic structures in \textit{Leaves of Grass}, see James Perrin Warren, ``Style and Technique(s),'' in \textit{The Routledge Encyclopedia of Walt Whitman}, eds. J.R. LeMaster and Donald D. Kummings (New York: Routledge, 1998), 693-696.} It is therefore meaningful to study how the stanzas vary statistically in different editions as it reveals how poetic structures change in \textit{Leaves of Grass} with respect to time, which could shed lights on potential stylistic shift. In this section, all stanzas (a poem with only one stanza is treated as one stanza) are extracted, and statistical tools are applied to computing the average as well as the standard deviation of lengths of stanzas in all editions.

\begin{table}[h!]
\begin{tabular}{c|c|c}
\hline
Edition & Mean (by \# of lines) & Standard Deviation (by \# of lines) \\
\hline
1855 & 3.60 & 5.37\\
\hline
1856 & 4.39 & 6.04\\
\hline
1860-61 & 4.75 & 6.54\\
\hline
1867 & 4.87 & 5.91\\
\hline
1871-72 & 5.01 & 6.33\\
\hline
1881-82 & 5.10 & 5.40\\
\hline
1891-92 & 5.20 & 5.33\\
\hline
\end{tabular}
\caption{Stanza size and variation in \textit{Leaves of Grass}}
\label{tab::stanzaSizes}
\end{table}

From Table \ref{tab::stanzaSizes}, it is obvious that the average stanza size kept growing as Whitman updated his \textit{Leaves of Grass}. Whitman did become wordy when he grew old in this sense, which could undermine the quality of poems written in his late career.

\par
The standard deviation first increases to over 6, then drops to around 5.3 in the last edition. We observe a decrease in the standard deviation of stanza size in the last decade of Whitman's writing career, which indicates that Whitman engaged in less artistic manipulation of stanza forms in his last years. As some literary critics suggest, the decrease in irregularity of stanzas suggests a waning of Whitman's poetic power.\footnote{James Perrin Warren, ``Style and Technique(s),'' in \textit{The Routledge Encyclopedia of Walt Whitman}, eds. J.R. LeMaster and Donald D. Kummings (New York: Routledge, 1998), 693-696.}

\section{Word/Punctuation Frequencies}
It is often the case that the use of certain words suggests one writer's language style. The formula for computing the frequency of a certain word $\omega$ in a text is
\[\text{Frequency of }\omega=\frac{\text{Sum of Appearances of } \omega}{\text{Total Number of Words}}.\]
When reading \textit{Leaves of Grass}, we can always feel Whitman's pride in his self-identity. For example, the first edition (1855) of \textit{Leaves of Grass} starts its first line with the celebration of Whitman himself:
\begin{verse}
I CELEBRATE myself, \\
And what I assume you shall assume, \\
For every atom belonging to me as good belongs to you.\footnote{Walt Whitman, \textit{Leaves of Grass}, 1855, The Walt Whitman Archive, 13.} \\
\end{verse}
In this short stanza, words referring to Whitman's self-identity appear for 4 times in total: 2 ``I''s, 1 ``myself'', and 1 ``me''. If we count the frequencies of words referring to oneself in all editions, we can portray how Whitman's self-identity changes throughout his career.
\begin{table}[h!]
\centering
\begin{tabular}{c|c|c|c}
\hline
Edition & ``I'' (\%) & ``me/my/mine/myself/self'' (\%) & Sum \\
\hline
1855 & 1.98 & 1.61 & 3.59\\
\hline
1856 & 2.08 & 1.40 & 3.48\\
\hline
1860-61 & 2.23 & 1.65 & 3.88\\
\hline
1867 & 2.16 & 1.66 & 3.83\\
\hline
1871-72 & 2.09 & 1.60 & 3.69 \\
\hline
1881-82 & 2.07 & 1.56 & 3.63\\
\hline
1891-92 & 1.97 & 1.51 & 3.48\\
\hline
\end{tabular}
\caption{Frequencies of self-identity words in \textit{Leave of Grass}}
\label{tab::self}
\end{table}

Table \ref{tab::self} lists frequencies of self-identity words in different editions. For example, the word ``I'' composes 1.98 percent of all words in the 1855 edition. Note that the results are not case sensitive, which means both ``my'' and ``My'' are counted, and so on.

\par
As shown in Table \ref{tab::self}, Whitman used fewer and fewer self-identity words starting from the 1867 edition, the first edition published after the Civil War. Possibly the trauma of the cruel war made Whitman start questioning the meaning of ``self''. So many people died as Whitman witnessed it in Washington DC.. Is one's self still worth celebrating if everyone is going to die someday?  This kind of pessimistic mindset is best shown in ``Year that Trembled and Reel'd beneath Me'', a poem first appearing in the 1867 edition:
\begin{verse}
YEAR that trembled and reel'd beneath me! \\
Your summer wind was warm enough—yet the air I
breathed froze me; \\
A thick gloom fell through the sunshine and darken'd
me; \\
Must I change my triumphant songs? said I to myself; \\
Must I indeed learn to chant the cold dirges of the baffled? \\
And sullen hymns of defeat?\footnote{Walt Whitman, \textit{Leaves of Grass}, 1867, The Walt Whitman Archive, 54.}\\
\end{verse}
``Must I change my triumphant songs? said I to my self;'' is in strong contrast to the affirmative and confident tone in ``I CELEBRATE myself'' previously in the first edition (1855), conveying a doubtful attitude to oneself. In some sense the Civil War reshaped Whitman's self-identity. 

\par
Moreover, the frequencies of self-identity words did not bounce back in later editions as the trauma of the Civil War gradually diminished. This result is opposite to that of sentiment analysis discussed later, which does rebound in later editions. There must be something deeper causing the strange pattern of Whitman's self-identity. Ed Folsom claims that the ``I'' we expect to appear in \textit{Leaves of Grass} is buried in the homophonic ``aye'' after analyzing the ``Annexes'' in detail in the ``deathbed'' edition. Whitman in his last years ``wrote poems that track a self that is diffusing back into the world it emerged from, observing its physical body decaying, tracking its wandering thoughts, noting the things that are now beginning to exist not as a part of the self but as a part of the world the self will no longer be absorbing. `I' is swallowed up in the eternity of `aye'.''\footnote{Ed Folsom, ``Whitman Getting Old,'' in \textit{The New Walt Whitman Studies}, ed. Matt Cohen (Cambridge: Cambridge University Press 2019), 232-247.}

\par
In addition to self-identity words, the use of parentheses in \textit{Leaves of Grass} has drawn attention from Whitman scholars. In fact, there was a dramatic change in the use of parentheses during Whitman's career. In the first edition (1855) of \textit{Leaves of Grass}, only two sets of parentheses appear in the entire text. Here we list frequencies of parentheses in different editions in Table \ref{tab::parentheses}.
\begin{table}[h!]
\centering
\begin{tabular}{c|c}
\hline
Edition & Parentheses (\textperthousand) \\
\hline
1855 & 0.0538\\
\hline
1856 & 0.309\\
\hline
1860-61 & 1.15 \\
\hline
1867 & 2.29\\
\hline
1871-72 & 2.67\\
\hline
1881-82 & 2.95\\
\hline
1891-92 & 3.30\\
\hline
\end{tabular}
\caption{Frequencies of parentheses in \textit{Leaves of Grass}}
\label{tab::parentheses}
\end{table}

\par
There is an increase of the use of parentheses in each edition, especially in the 1860-61 edition and 1867 edition. Literary scholars tend to attribute the drastic increase in parentheses in the 1867 edition to the postwar trauma. The idea is that Whitman treated the Civil War itself as ``a parenthetical moment--a break from the normalcy of the national history--and a clarifying realization of American purpose and ideals.''\footnote{Kenneth M. Price, ``Love, War, and Revision in Whitman's Blue Book,'' Huntington Library Quaterly, 73.4 (2010), 679-692.} 

\par
Another interpretation of parentheses is that they make texts fragile. During the war Whitman had the chance to travel to the battlefield in search of his brother, who enlisted in the Union army when the war started. At a field hospital in Fredericksburg, he encountered a sight that he would never forget: "$\dots$a heap of amputated feet, legs, arms, hands, etc., a full load for a one-horse cart." As he wrote in the journal: "$\dots$human fragments, cut, bloody, black and blue, swelled and sickening."\footnote{Ed Folsom and Kenneth M. Price, \textit{Re-Scripting Walt Whitman: An introduction to his life and work} (Malden: Blackwell Publishing, 2005), 79-80.} Whitman had long been a poet who celebrated physical body. For example, Whitman celebrated himself for ``every atom'' belonging to him at the beginning of the first edition (1855).\footnote{Walt Whitman, \textit{Leaves of Grass}, 1855, The Walt Whitman Archive, 13.} The horrible scene of fragile human bodies due to the war, resembling fragile texts caused by parentheses, was a shock to Whitman, and as a result, he reconstructed his Leaves of Grass, along with the reconstruction of the nation, by integrating the trauma of the Civil War (symbolized by parentheses) with the original hope for a democratic future for America.

\par
Aside from the 1867 edition, the increasing use of parentheses in the 1860-61 edition reveals a less-known fact that the anticipation of the war, very much in the air at the time, began to affect Whitman's use of punctuation earlier--that is, that a focus on trauma as a result of the war might be turned instead to the broader emotional category of anxiety, and thus decoupled from the particular events of the war.

\section{Macro-Etymological Analysis and Lexical Complexity}
Aside from studying frequencies of featured words, it is also worthwhile to study the origins of words, i.e. etymological analysis. As a language with various etymological sources, English is rich in synonyms, and people can convey one thought in multiple ways. Therefore, a macro-etymological analysis here could help us understand Whitman's etymological preference and how it evolved throughout his life. 

\par
The etymological distribution of words in \textit{Leaves of Grass} is shown in Table \ref{tab::etym}. The results are obtained with a macro-etymological analyzer based on the Etymological Wordnet.\footnote{For a detailed description of the macro-etymological analyzer, see Jonathan Reeve, ``The Macro-Etymological Analyzer,'' accessed August 25, 2021, \url{https://github.com/JonathanReeve/macro-etym/}.} Note that some minor etymological sources of \textit{Leaves of Grass} (e.g. Italian, Ancient Greek, etc.) are not listed.

\begin{table}[h!]
\centering
\begin{tabular}{c|c|c|c|c|c}
\hline
Edition & Old English & Old French & Latin & French & Anglo- \\
 & (ca. 450-1100) & (842-ca. 1400) & & &Norman \\
\hline
1855 & 41.95 & 20.59 & 13.98 & 6.50 & 6.51\\
\hline
1856 & 38.33 & 21.24 & 15.93 & 7.18 & 6.42\\
\hline
1860-61 & 36.12 & 21.14 & 17.88 & 7.45 & 6.72\\
\hline
1867 & 36.04 & 20.79 & 17.92 & 7.92 & 6.50\\
\hline
1871-72 & 36.11 & 20.92 & 17.62 & 7.86 & 6.46\\
\hline
1881-82 & 34.71 & 21.07 & 18.96 & 7.87 & 6.37\\
\hline
1891-92 & 34.47 & 20.93 & 19.28 & 8.08 & 6.27\\
\hline
\end{tabular}
\caption{Etymological distribution of words in \textit{Leaves of Grass}}
\label{tab::etym}
\end{table}

\par
As shown in Table \ref{tab::etym}, the use of words from French increases in almost each edition of Leaves of Grass except for a small decline in the 1871-72 edition, suggesting Whitman's favor of French words. In the last edition (1891-92) around 8\% of words in \textit{Leaves of Grass} are from French.

\par
Whitman's passion for French words could be explained from the perspective of his personal experience prior to the writing of \textit{Leaves of Grass}. Whitman left the New York area for the first time in his life in 1848, heading for New Orleans with his younger brother. He encountered the diversity of America for the first time there, a city that fit him more than New York. It is highly possible that Whitman developed his fondness for using French words during his short stay in the south.\footnote{Ed Folsom and Kenneth M. Price, \textit{Re-Scripting Walt Whitman: An introduction to his life and work} (Malden: Blackwell Publishing, 2005), 13.} In the preface to the first edition, Whitman expressed his praise for French: ``The English language befriends the grand American expression .... it is brawny enough and limber and full enough. On the tough stock of a race who through all change of circumstance was never without the idea of political liberty, which is the animus of all liberty, it has attracted the terms of daintier and gayer and subtler and more elegant tongues.''\footnote{Walt Whitman, \textit{Leaves of Grass}, 1855, The Walt Whitman Archive, xi-xii.} The ``more elegant tongues'' here refer to French, Latin, and Greek.\footnote{James Perrin Warren, \textit{Walt Whitman's Language Experiment} (University Park: Penn State University Press, 1990), 35.}

\par
From Table \ref{tab::etym}, the percentage of words originating from Latin, another language in the ``more elegant tongues'' by Whitman, follows the same pattern. Latinate words increase in every edition except for the 1871-72 edition, peaking at nearly 20\% of total words in the 1891-92 edition. A study compared Whitman's proses during his late years to his early poetry and found the same ``entrapment in an elaborate, nonconversational, heavily Latinate, intentionally complicated style'' with his late poems.\footnote{C. Carrol Hollis, \textit{Language and Style in Leaves of Grass} (Baton Rouge: Louisiana State University Press, 1983), 224.} The author inferred that the real cause of the stylistic shift was the Harlan dismissal in 1865, when James Harlan, Secretary of the Interior, discharged Whitman from his second-class clerkship in the Bureau of Indian Affairs because he happened to read the drafts for subsequent \textit{Leaves of Grass} in Whitman's office and declared the book obscene and its author immoral.\footnote{Joseph P. Hammond, ``Harlan, James W. (1820-1899),'' in \textit{The Routledge Encyclopedia of Walt Whitman}, eds. J.R. LeMaster and Donald D. Kummings (New York: Routledge, 1998), 265-266.} To save Whitman's reputation, his close friend and sponsor O'Connor wrote a long article named ``Good Gray Poet'', transforming Whitman from a young radical into a kind and patriotic poet in the public eye.\footnote{Ed Folsom, ``Whitman Getting Old,'' in \textit{The New Walt Whitman Studies}, ed. Matt Cohen (Cambridge: Cambridge University Press, 2019), 232-247.} Whitman was then under great pressure to live up to what he was expected to be after this rebranding process, which forced him to get rid of his early style and turned to the intentionally complicated style instead.\footnote{C. Carrol Hollis, \textit{Language and Style in Leaves of Grass} (Baton Rouge: Louisiana State University Press, 1983), 224.} A more detailed tabulation of Latinate words in Whitman's early and late poetry can be found in the book by C. Carroll Hollis.\footnote{C. Carrol Hollis, \textit{Language and Style in Leaves of Grass} (Baton Rouge: Louisiana State University Press, 1983), 255-256.} There is also a study on the frequently used Latinate words in \textit{Leaves of Grass}.\footnote{Jonathan Reeve, ``A Comparative Macro-Etymology of Whitman Editions,'' accessed August 25, \url{https://jonreeve.com/2014/09/macroetymology-of-whitman-editions/}.}

\par
The heavy use of Latinate words and the ``intentionally complicated style'' in late editions could lead to an increase in the lexical complexity/richness in Whitman's works as Latinate words are generally longer and multisyllabic. For the general public, a text is more difficult to comprehend if Germanic words are replaced with their Latinate equivalents.

\par
Lexical complexity can be measured in many different ways, which are mainly categorized into three classes: lexical density, lexical sophistication, and lexical variation.\footnote{John Read, \textit{Assessing Vocabulary} (Cambridge: Cambridge University Press, 2000), 200-201.}

\par
Lexical density (LD) of a document is the proportion of lexical words, i.e.,
\[\text{LD}=\frac{\text{Number of Lexical Words}}{\text{Total Number of Words}}.\]
Lexical words in a text are those that convey meanings, as opposed to function words, which exist only for grammatical correctness. In this study, lexical words consist of nouns, adjectives, verbs, and adverbs derived from adjectives.\footnote{John Read, \textit{Assessing Vocabulary} (Cambridge: Cambridge University Press, 2000), 203.}

\par
Lexical sophistication of a document is the proportion of ``unusual and or advanced words''.\footnote{John Read, \textit{Assessing Vocabulary} (Cambridge: Cambridge University Press, 2000), 203.} The corresponding measure of lexical sophistication (LS) is defined as
\[\text{LS}=\frac{\text{Number of Sophiscated Word Types}}{\text{Total Number of Word Types}}.\]
By definition, a word type of a text is a word that appears for at least once in this text. In this study a word is ``sophisticated'' if it is not among the 2,000 most frequent words from the British National Corpus.\footnote{Geoffrey Leech, Paul Rayson, and Andrew Wilson, \textit{Word Frequencies in Written and Spoken English: Based on the British National Corpus} (New York: Routledge, 2014), 120-125.}

\par
Traditionally the measure of lexical variation is the type-token ratio (TTR)
\[\text{TTR}=\frac{\text{Total Number of Word Types}}{\text{Total Number of Words}}.\]
However, studies have found that TTR depends on the text length. As the text length increases, the TTR decreases.\footnote{Brian Richards, ``Type/Token Ratios: what do they really tell us?'' Journal of Child Language, 14.2 (1987), 201-209.} Given that \textit{Leaves of Grass} expanded quickly as Whitman added new poems, TTR is not an appropriate measure of lexical variation in this case. A simple calculation shows that the TTR value of \textit{Leaves of Grass} drops from 0.185 (the 1855 edition) to 0.106 (the 1891-92 edition) due to an increase in the text length.

\par
To fix the problem of length dependence with TTR, researchers invented many new measures of lexical variation based on TTR. For example, the D measure (claimed to be independent of the text length) of lexical variation is obtained by solving the following equation
\[\text{TTR}=\frac{D}{N}\left[\left(1+2\frac{N}{D}\right)^{1/2}-1\right],\]
where $N$ is the total number of words and $D$ is the D measure of lexical variation.\footnote{Pilar Dur\'{a}n et al., ``Developmental Trends in Lexical Diversity,'' Applied Linguistics, 25.2 (2004), 220-242.}

\par
An alternative way to measure lexical variation is mean segmental TTR (MSTTR), which is the average of TTR values of successive segments of a text. For short texts, this method is problematic since most short texts cannot be divided into segments with equal lengths. For example, if a text consists of 120 words and the given length of a segment is 50 words, then the last segment consists of only 20 words, undermining the validity of MSTTR. For long texts such as \textit{Leaves of Grass}, MSTTR is a fairly good measure for lexical variation. In this study we calculate MSTTR by dividing each edition of \textit{Leaves of Grass} into segments of 50 words (approximately the length of a typical stanza in \textit{Leaves of Grass}).

The results of lexical complexity analysis are shown in Table \ref{tab::complexity}. The texts are first tokenized and tagged using the word tokenizer and the part of speech tagger in \verb|nltk|, a python package for natural language processing tasks. The output is then lemmatized with the Wordnet lemmatizer in \verb|nltk|, which ensures that we count words in different forms as a single word type (e.g. ``come'', ``came'', ``coming'', and ``comes'') in lexical analysis.\footnote{Steven Bird, Ewan Klein, and Edward Loper, \textit{Natural Language Processing with Python} (Sebastopol: O'Reilly Media, 2009), 179-189.} The final step is to apply a lexical complexity analyzer to the texts already processed.\footnote{The powerful lexical complexity analyzer gives nearly 30 different measures of lexical complexity, see Xiaofei Lu, ``The Relationship of Lexical Richness to the Quality of ESL Learners’ Oral Narratives,'' The Modern Language Journal, 96.2 (2012), 190-208.}
\begin{table}[h!]
\centering
\begin{tabular}{c|c|c|c}
\hline
Edition & LD & LS & MSTTR \\
\hline
1855 & 0.489 & 0.806 & 0.717\\
\hline
1856 & 0.510 & 0.859 & 0.723\\
\hline
1860-61 & 0.507 & 0.881 & 0.737\\
\hline
1867 & 0.510 & 0.888 & 0.741\\
\hline
1871-72 & 0.512 & 0.881 & 0.742\\
\hline
1881-82 & 0.516 & 0.883 & 0.742\\
\hline
1891-92 & 0.519 & 0.889 & 0.750\\
\hline
\end{tabular}
\caption{Lexical complexity in \textit{Leaves of Grass}}
\label{tab::complexity}
\end{table}

\par
As shown in Table \ref{tab::complexity}, all of the three components of lexical complexity increase from the 1855 edition to the 1891-92 edition, meeting our expectation that the rise of Latinate words leads to greater lexical complexity.

\section{Principal Component Analysis}
Stylometry is about extracting featured words of texts, and comparing them with each other using statistical methods. In this section the principal component analysis (PCA) is used to visualize the distribution of styles of different editions in 2 dimensions.

\par
The first step is to extract the featured words among all editions. Since we have did analysis of certain style words (e.g. self-identity words) in sections above, this section is focused on featured content words only. Importing the stopwords from \verb|nltk|, we are left with only content words for subsequent featured words extraction.\footnote{For a sample list of stopwords, see Steven Bird, Ewan Klein, and Edward Loper, \textit{Natural Language Processing with Python} (Sebastopol: O'Reilly Media, 2009), 60-61.}

\par
Next we use the term frequency–inverse document frequency (tf-idf) method to rank the importance of words across all documents considered here. The tf-idf score for a word $\omega$ in a document $d$ is computed by multiplying the tf score by the idf score
\[\text{tf-idf}(d, \omega) = \text{tf}(d, \omega)\cdot \text{idf}(d, \omega).\]
By default,\footnote{For the workflow of tf-idf, see ``Scikit-Learn User Guide'', accessed August 25, 2021, \url{https://scikit-learn.org/stable/modules/feature\_extraction.html}.} the tf score $\text{tf}(d, \omega)$ is the number of appearances of the word $\omega$ in document $d$. The idf score is defined as
\[\text{idf}(d, \omega) = 1+\ln\frac{1+n}{1+\text{df}(\omega)},\]
where $n$ is the total number of documents in the corpus and $\text{df}(\omega)$ is the number of documents in the corpus that contain the word $\omega$. The $\text{tf-idf}$ scores are then normalized such that the Euclidean norm of the vector for each document is 1, i.e.
\[\text{tf-idf}(d,\omega)\longrightarrow \frac{\text{tf-idf}(d,\omega)}{\sqrt{\sum_{\omega_i \in d}\text{tf-idf}(d,\omega_i)^2}}.\]
After computing tf-idf scores of all words across 7 editions of \textit{Leaves of Grass}, we select the most featured 400 words (i.e. those with highest tf-idf scores) for subsequent PCA.

\par
\begin{figure}[h!]
\centering
\includegraphics[width=0.8\textwidth]{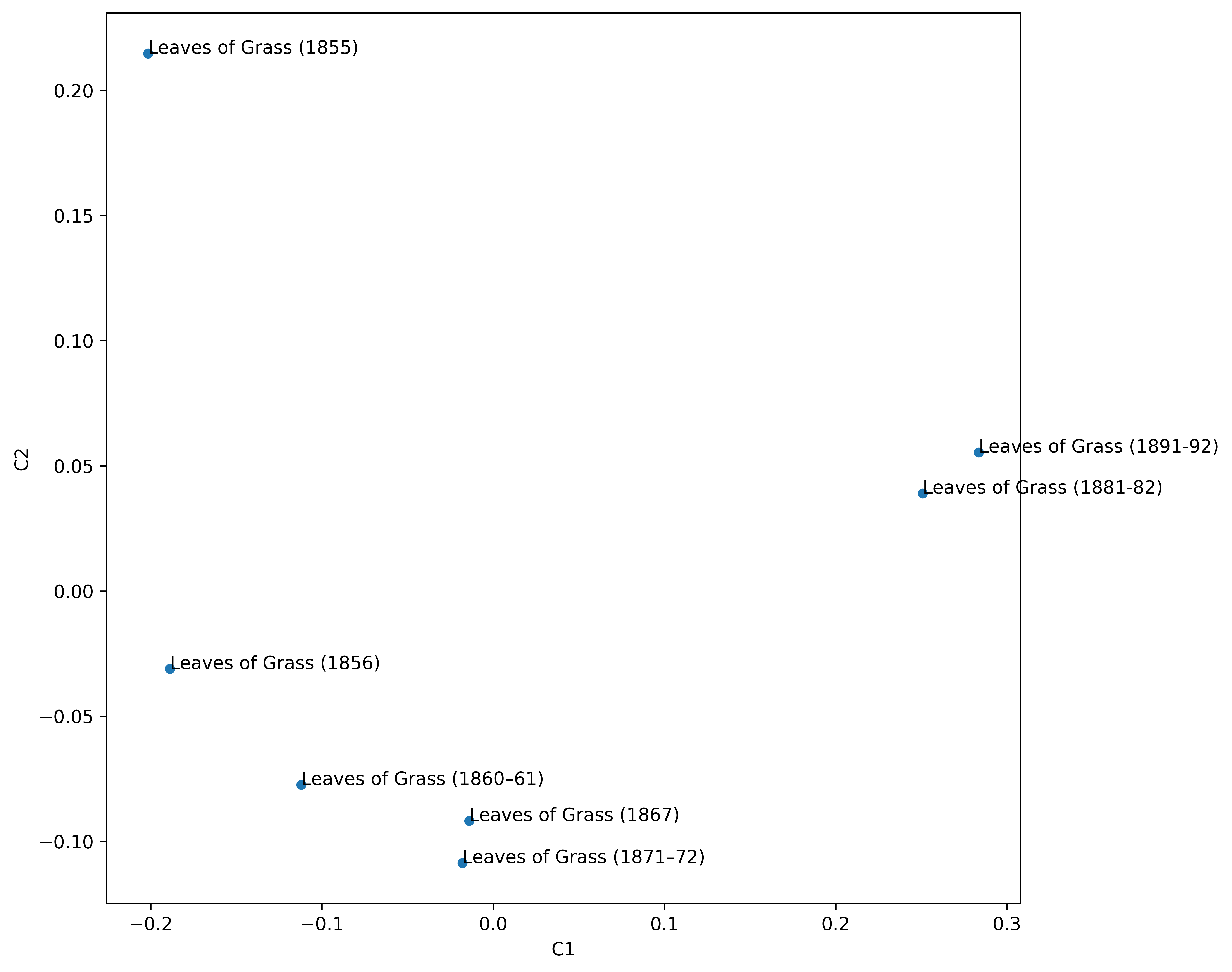}
\caption{2-dimensional PCA of featured words in \textit{Leaves of Grass}}
\label{fig::PCA}
\end{figure}

PCA finally reduces the dimensionality from 400 to 2 while still preserving distances among points to the maximum degree possible, and hence we can visualize the differences of contents among texts by evaluating distances among points in the PCA plot. The results are shown in Figure \ref{fig::PCA}.

\par
As we can see in Figure \ref{fig::PCA}, the stylistic difference between the 1855 edition and the 1856 edition is huge. The 1856 edition was the first major revision Whitman made to \textit{Leaves of Grass} since its debut, and Whitman was still experimenting with various techniques to form his own poetic style. Consequently, the 1856 edition is very different from its predecessor. The second largest difference appears between the 1881-82 edition and the 1871-72 edition. As shown in Table \ref{tab::stanzas}, the 1881-82 edition is the single largest expansion made by Whitman as the number of stanzas increases from 1381 to 1948. In fact, Whitman viewed the 1881-82 edition as the definitive edition, the unification of his previous works.\footnote{Dennis K. Renner, ``Leaves of Grass, 1881-82 edition,'' in \textit{The Routledge Encyclopedia of Walt Whitman}, eds. J.R. LeMaster and Donald K. Kummings (New York: Routledge, 1998), 373-375.} Moreover, there is a 10 year gap between the 1881-82 edition and the 1871-72 edition, during which Whitman certainly changed his style a bit. It is therefore not surprising to see such a big stylistic difference. There is also a 10 year gap between the 1891-92 edition and the 1881-82 edition, but Whitman did nothing but add two short ``Annexes'' in this period, leading to a fairly small stylistic shift in Figure \ref{fig::PCA}.

\par
\begin{figure}[h!]
\centering
\includegraphics[width=0.8\textwidth]{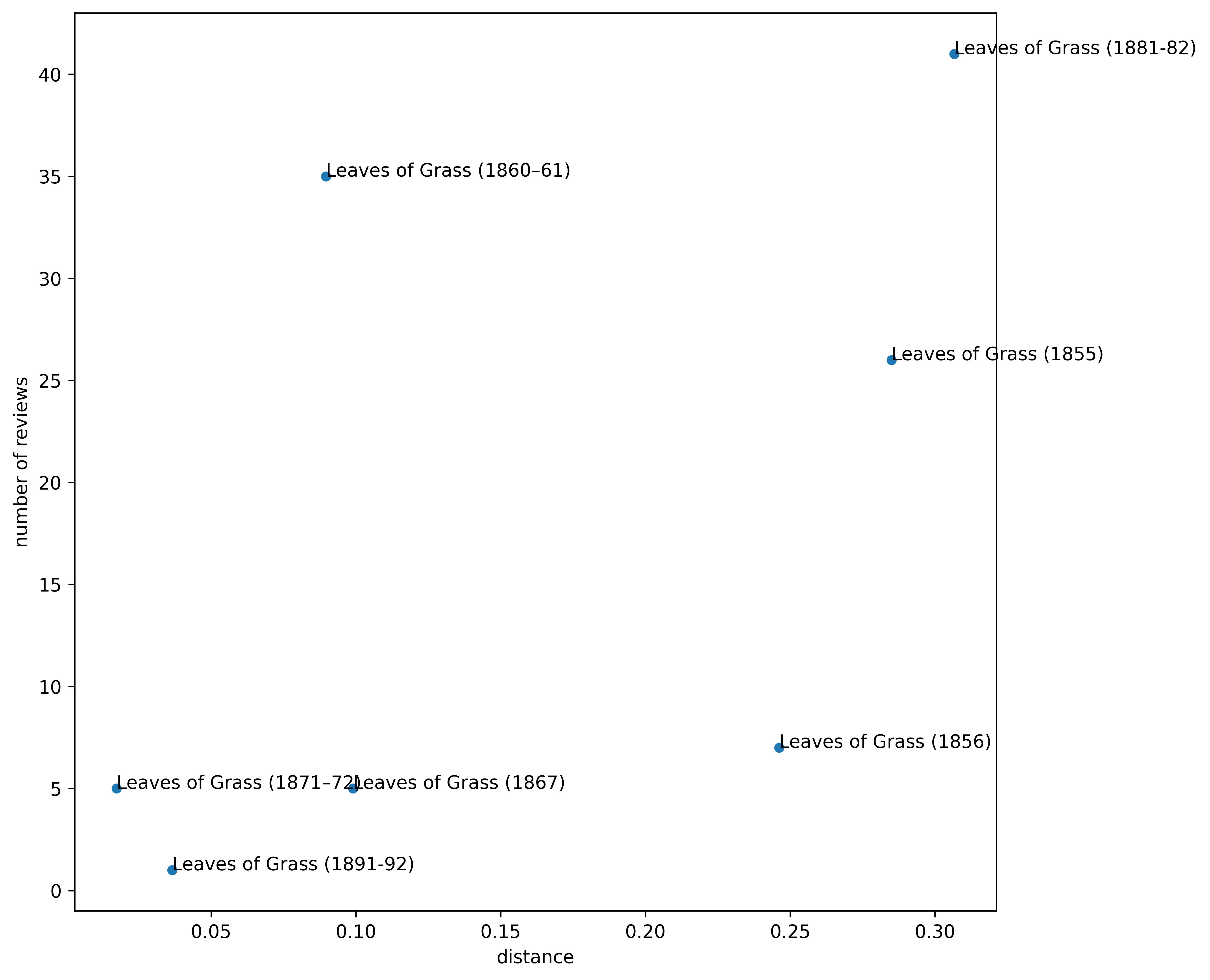}
\caption{Number of reviews vs. distances for \textit{Leaves of Grass}}
\label{fig::correlation}
\end{figure}

To further justify the use of PCA, we collect the number of review articles on each edition at the time of publication available on the Whitman Archive, as well as the Euclidean distance between each edition from the previous one in Figure \ref{fig::PCA}.\footnote{The review articles can be viewed on the following site: Matt Cohen, Ed Folsom, and Kenneth M. Price, ``Commentary: Contemporary Reviews'', accessed August 25, 2021, \url{https://whitmanarchive.org/criticism/reviews/}.} We expect that the more distinct some edition is from its predecessor, the more attention it receives from the media, and therefore the more review articles available at the time of publication. As shown in Figure \ref{fig::correlation}, a fairly good correlation exists between the two quantities, with the Pearson correlation coefficient (one-tailed)
\[r=0.579, \, p=0.0866.\]

\par
Whitman is well-recognized as a poet who cared a lot about his reception. In a recent study of his correspondence, researchers have found that Whitman wrote many letters to people around the world who could promote his works. Whitman even gave copies of his works for free along with these letters to his advocates. In some sense ``Whitman was an expert of the social media of his day, and his promotional effort seems to have paid off...''\footnote{Alexander Ashland, Stefan Sch\"{o}berlein, and Stephanie M. Blalock, ``Reading Whitman's Epistolary Database,'' in \textit{The New Walt Whitman Studies}, ed. Matt Cohen (Cambridge: Cambridge University Press, 2019), 121-143.}

\par
In fact, Whitman disliked the critics in his early career since most of them were negative. One 1856 review even suggested Whitman be sent to an insane asylum. To combat these negative reviews and bolster sales, Whitman even wrote favorable reviews of his own work. One example is a short piece appearing in the United States Review in 1855, praising Whitman himself as ``an American bard at last!''\footnote{The original review written by Whitman himself is in Walt Whitman, ``Walt Whitman and His Poems,'' The United States Review 5 (September 1855), 205-12. This piece is also archived in The Walt Whitman Archive. The link is \url{https://whitmanarchive.org/criticism/reviews/leaves1855/anc.00176.html}.} Gradually, as his works became more and more recognized by the public, he started to receive more favorable reviews.\footnote{Shaunacy Ferro, ``When Walt Whitman Reviewed His Own Book,'' accessed August 25, 2021,\\ \url{https://www.mentalfloss.com/article/77096/when-walt-whitman-reviewed-his-own-book}.} Due to Whitman's dislike for early reviews on his work, a strict correlation between the style change and the number of reviews is not expected in the early editions. The two outliers in Figure \ref{fig::correlation} are the 1856 edition and the 1860-61 edition, both published in his early career. If we remove these two data points from Figure \ref{fig::correlation}, the Pearson correlation coefficient becomes $r=0.946, \, p=0.00747$, giving a very strong positive correlation between distance and number of reviews. In this respect, the use of PCA is justified.

\section{Sentiment Analysis}
Whitman wrote \textit{Leaves of Grass} in celebration of American democracy and liberty, but switched gears to incorporate trauma after the war in later editions, as a period of poetic reconstruction. The 1867 edition is the first postwar edition, being the most carelessly printed and the most chaotic among all editions. Maybe Whitman made use of such chaos as a representation of chaos and despair the war had brought to America.\footnote{Ed Folsom and Kenneth M. Price, \textit{Re-Scripting Walt Whitman: An introduction to his life and work} (Malden: Blackwell Publishing, 2005), 98.} As an example of Whitman's misery at that time, the assassination of President Lincoln in 1865 made Whitman wrote the poem ``O Captain! My Captain!'', first appearing in the 1867 edition (the first postwar edition). The following is the first stanza of the poem:
\begin{verse}
\textbf{1}\\
O CAPTAIN! my captain! our fearful trip is done; (-0.584) \\
The ship has weather'd every rack, the prize we sought is
won; (0.791) \\
The port is near, the bells I hear, the people all exulting, (0.0)\\
While follow eyes the steady keel, the vessel grim and daring: (-0.296)\\
But O heart! heart! heart! (0.0)\\
Leave you not the little spot, (-0.052)\\
Where on the deck my captain lies. (-0.422)\\
Fallen cold and dead.\footnote{Walt Whitman, \textit{Leaves of Grass}, 1867, The Walt Whitman Archive, 13.} (-0.778)\\
\end{verse}
Sentiment analysis, as a method to measure the emotional aspect of texts, showing changes in Whitman's sentiments quantitatively. We apply a lexicon-based sentiment analyzer following the sentiment analysis model named VADER to each sentence of \textit{Leaves of Grass} and get an average polarity score for each edition. The higher the polarity score is, the more positive the sentiment is. The polarity score is in the range -1 (extreme negative sentiments) to +1 (extreme positive sentiments). The standardized thresholds of sentence classification as either positive, neutral, or negative are
\begin{enumerate}
\item positive sentiment: $\text{score}\geq +0.05$,
\item neutral sentiment: $-0.05\leq\text{score}\leq +0.05$,
\item negative sentiment: $\text{score}\leq -0.05$.\footnote{The detailed calibration is in C. J. Hutto, ``VADER-Sentiment-Analysis,'' accessed Aug 25, 2021, \url{https://github.com/cjhutto/vaderSentiment}.}
\end{enumerate}

\par
As a test case, we first apply the sentiment analysis to sentences in the first stanza of ``O Captain! My Captain!'' above. The polarity scores are listed in the parentheses to the right of each sentence. The average polarity score of this stanza is -0.168, which is classified as negative sentiment, fitting our expectation.

\par
The results of sentiment analysis are listed in Table \ref{tab::sentiment}. The first thing to note is that all editions have average polarity scores above $+0.05$, indicating the existence of positive sentiments across all editions. No matter how much Whitman suffered from the tragedy of war, his celebration of America never vanished in \textit{Leaves of Grass}.
\begin{table}[h!]
\centering
\begin{tabular}{c|c}
\hline
Edition & Average Polarity Score\\
\hline
1855 & .0887 \\
\hline
1856 & .0924\\
\hline
1860-61 & .0908\\
\hline
1867 & .0793\\
\hline
1871-72 & .0795\\
\hline
1881-82 & .0834\\
\hline
1891-92 & .0811\\
\hline
\end{tabular}
\caption{Average Polarity Scores of \textit{Leaves of Grass}}
\label{tab::sentiment}
\end{table}

Taking a closer look on the results, we find that all editions before the war have average polarity scores around 0.09, with a plunge of sentiments in the 1867 edition. This dramatic decrease in sentiments is best explained by Whitman's postwar trauma since the 1867 edition is the first postwar edition. The average polarity scores of editions afterwards increase slightly in Whitman's old years, but are still much lower than the pre-war editions. A recent study by Ed Folsom seems to unravel the secret behind the rebound of sentiments in Whitman's late years.\footnote{Ed Folsom, ``Whitman Getting Old,'' in \textit{The New Walt Whitman Studies}, ed. Matt Cohen (Cambridge: Cambridge University Press, 2019), 232-247.} Folsom claims that Whitman hid his poetic philosophy within the 1855 triple urge sexual chant:
\begin{verse}
... \\
Urge and urge and urge, \\
Always the procreant urge of the world.\footnote{Walt Whitman, \textit{Leaves of Grass}, 1855, The Walt Whitman Archive, 14.}
\end{verse}
The ``transegmental drift'' in linguistics attaches the ``d'' of ``and'' to the ``u'' of ``urge'', turning the original sentence into ``urge and dirge and urge''. The poetic philosophy, wrote by Folsom, is that ``every procreant urge ultimately results in a death-dirge as the body is recycled into compost that eventually will urge out new life again.'' From sections above we know Whitman chanted ``the cold dirges of the baffled'' in the 1867 edition. In his later years, he seemed to step again into the ``urge'' phase of the ``urge and dirge and urge'' cycle. Whitman's returning to the beginning of his career in terms of sentiments is indicated from some lines in the 2 ``Annexes'' Whitman attached to the 1891-92 edition:
\begin{verse}
... \\
The wild unrest, the snowy, curling caps—that inbound urge
and urge of waves, \\
Seeking the shores forever.\footnote{Walt Whitman, \textit{Leaves of Grass}, 1891-92, The Walt Whitman Archive, 385.}\\
\end{verse}

Below is another example of Whitman's ``Old Age's Lambent Peaks'' in the 2 ``Annexes'':
\begin{verse}
...\\
The calmer sight—the golden setting, clear and broad: \\
So much i' the atmosphere, the points of view, the situations
whence we scan, \\
Bro't out by them alone—so much (perhaps the best) unreck'd
before; \\
The lights indeed from them—old age's lambent peaks.\footnote{Walt Whitman, \textit{Leaves of Grass}, 1891-92, The Walt Whitman Archive, 404.}\\
\end{verse}

\par
\begin{figure}[h!]
\begin{subfigure}{0.45\textwidth}
\includegraphics[width=\textwidth]{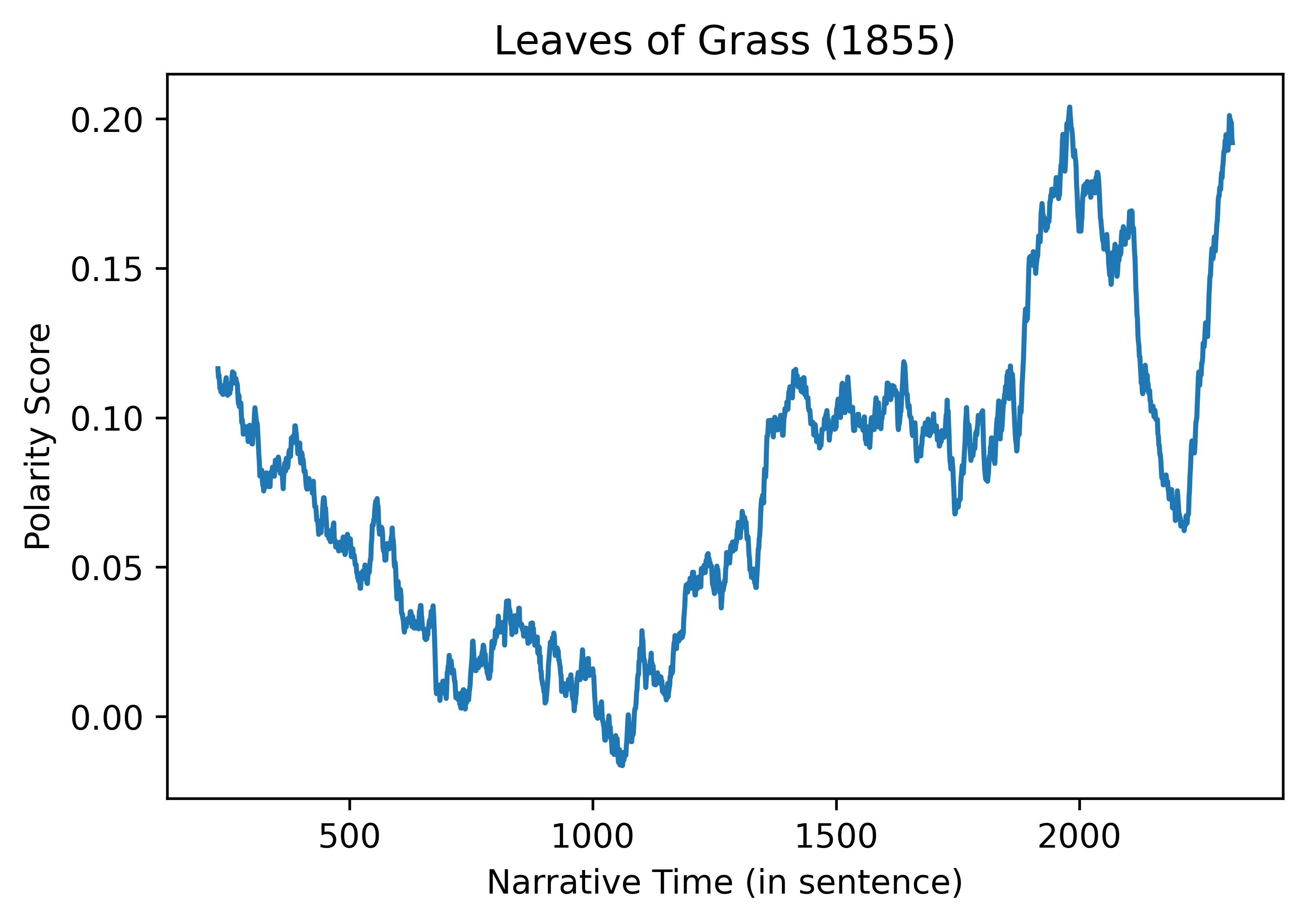}
\end{subfigure}
\begin{subfigure}{0.45\textwidth}
\includegraphics[width=\textwidth]{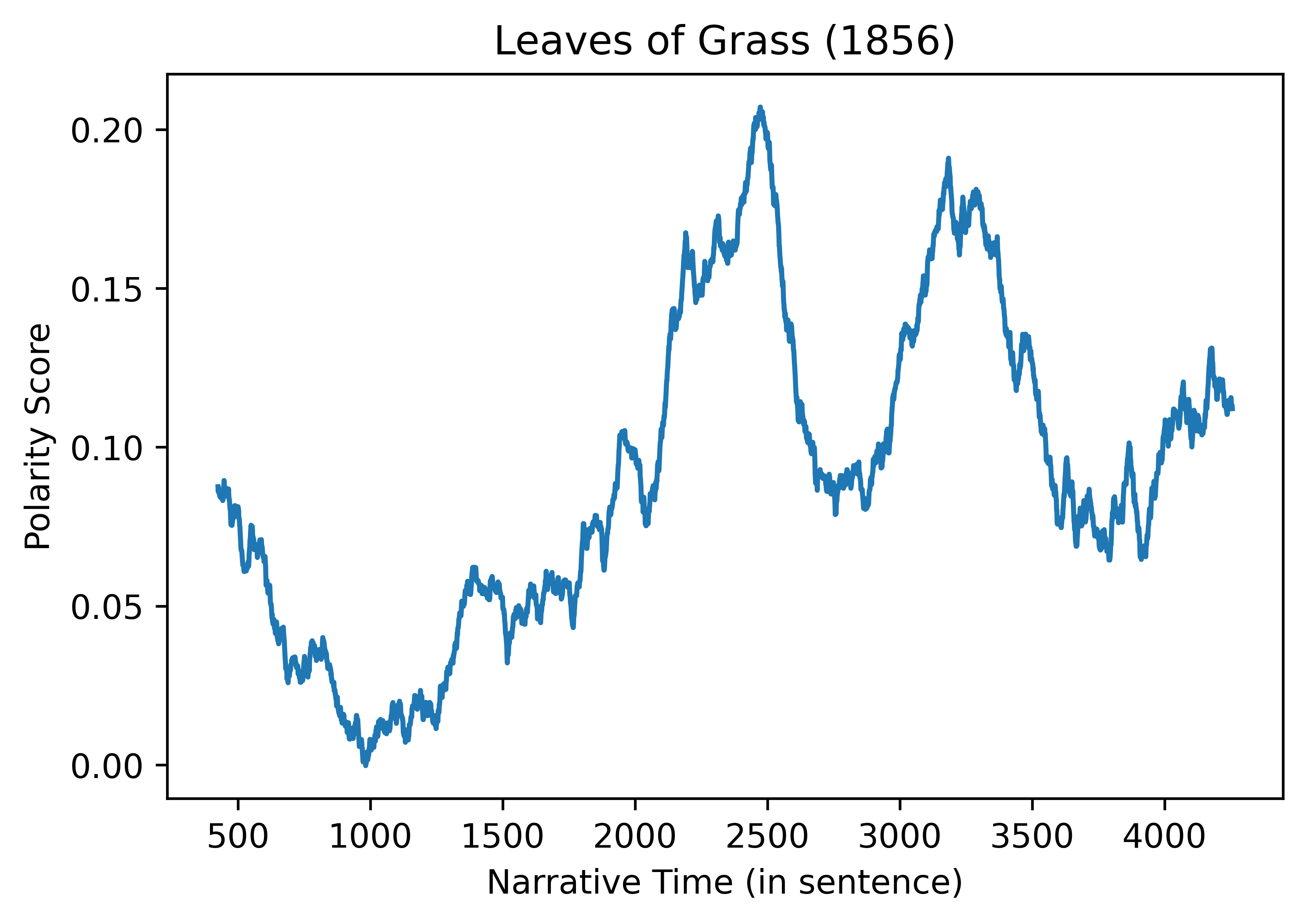}
\end{subfigure}
\begin{subfigure}{0.45\textwidth}
\includegraphics[width=\textwidth]{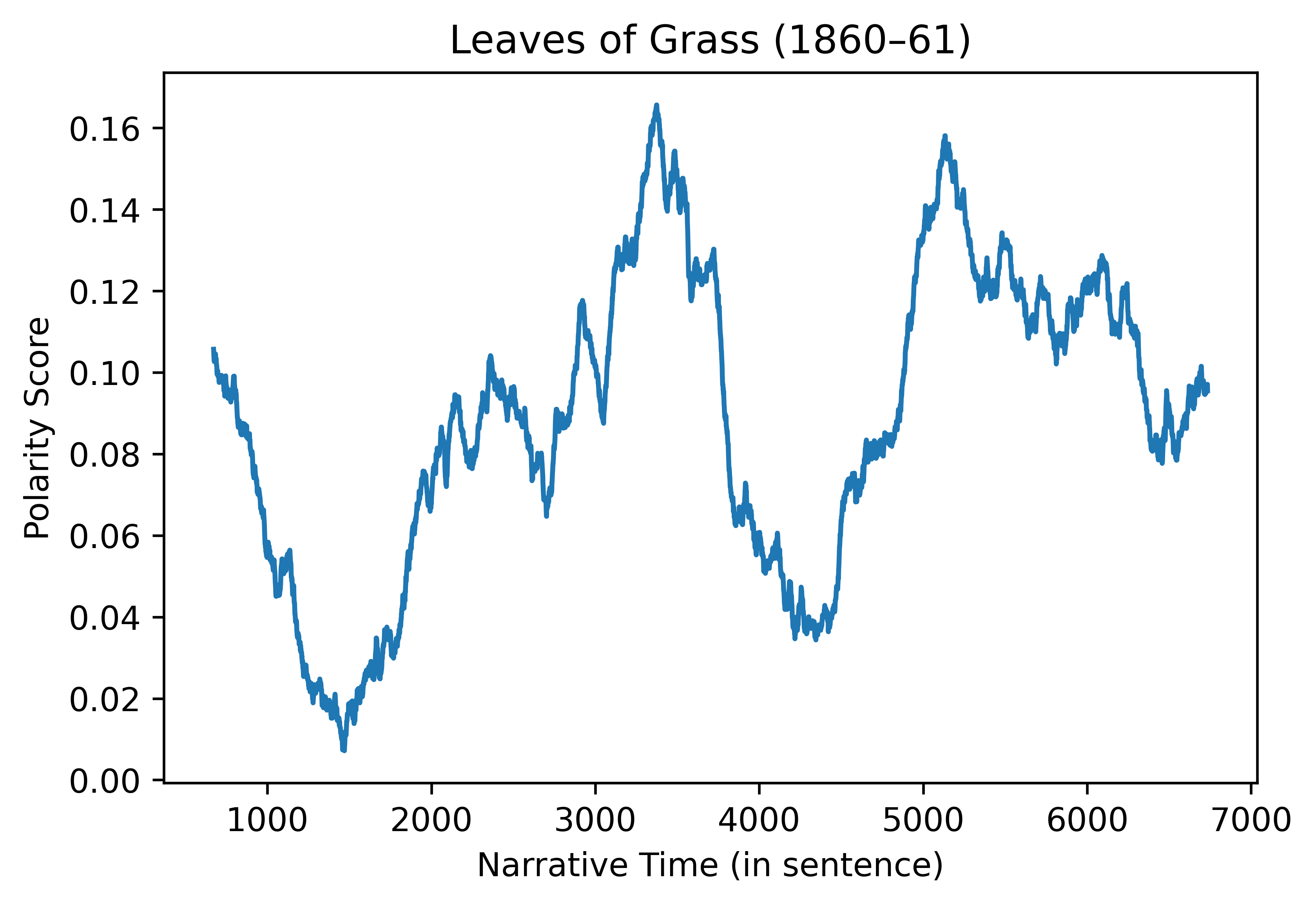}
\end{subfigure}
\begin{subfigure}{0.45\textwidth}
\includegraphics[width=\textwidth]{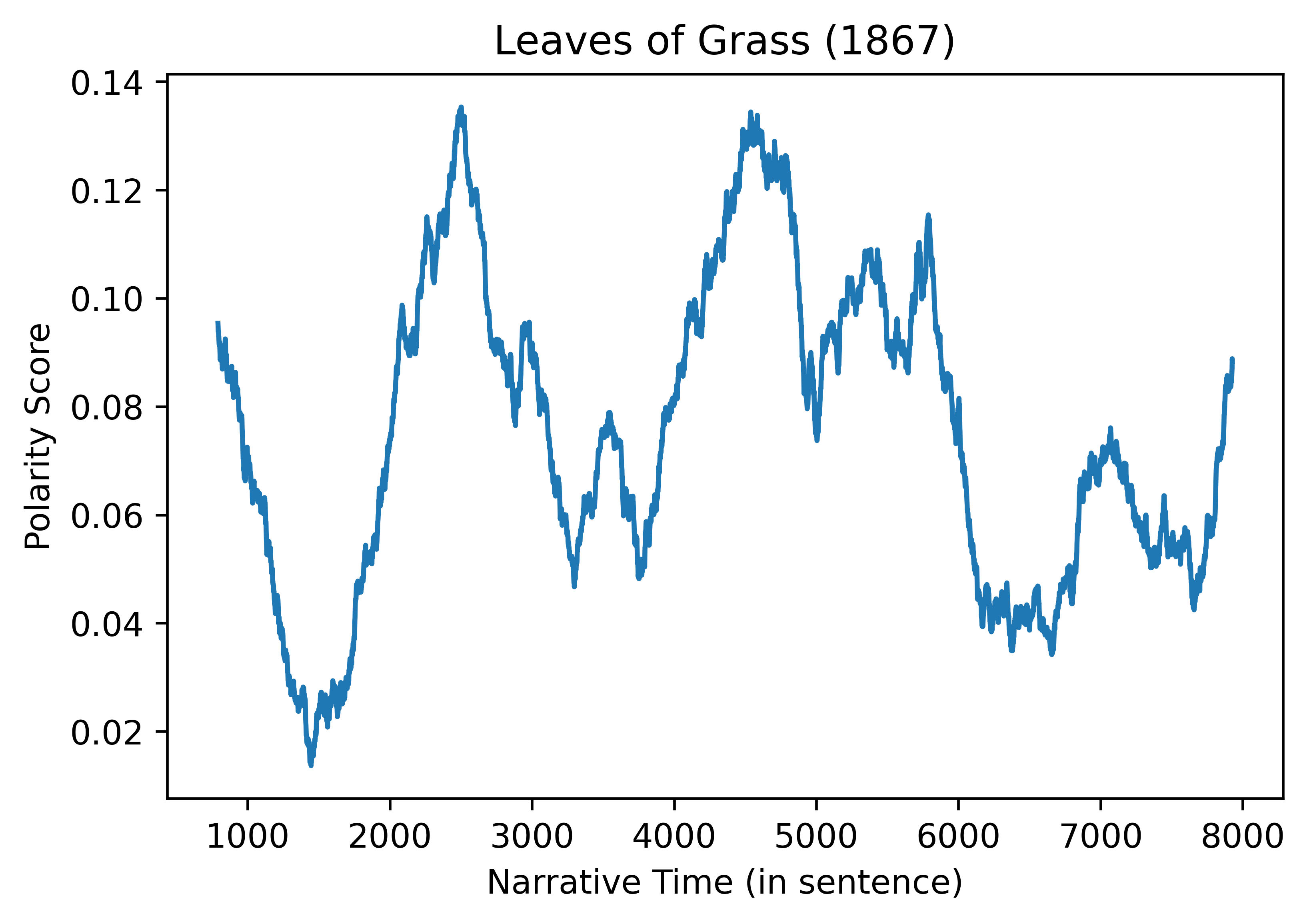}
\end{subfigure}
\begin{subfigure}{0.45\textwidth}
\includegraphics[width=\textwidth]{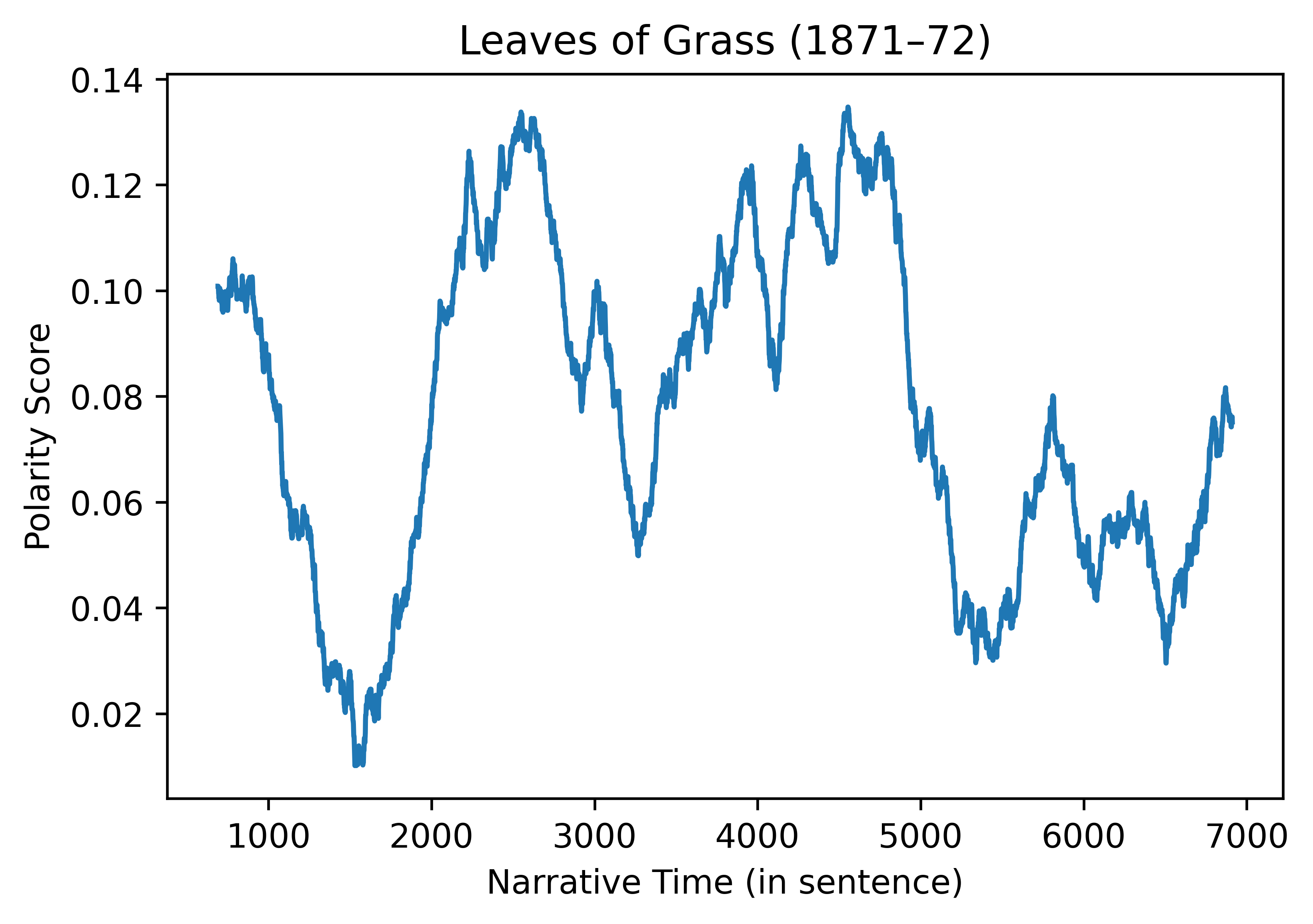}
\end{subfigure}
\begin{subfigure}{0.45\textwidth}
\includegraphics[width=\textwidth]{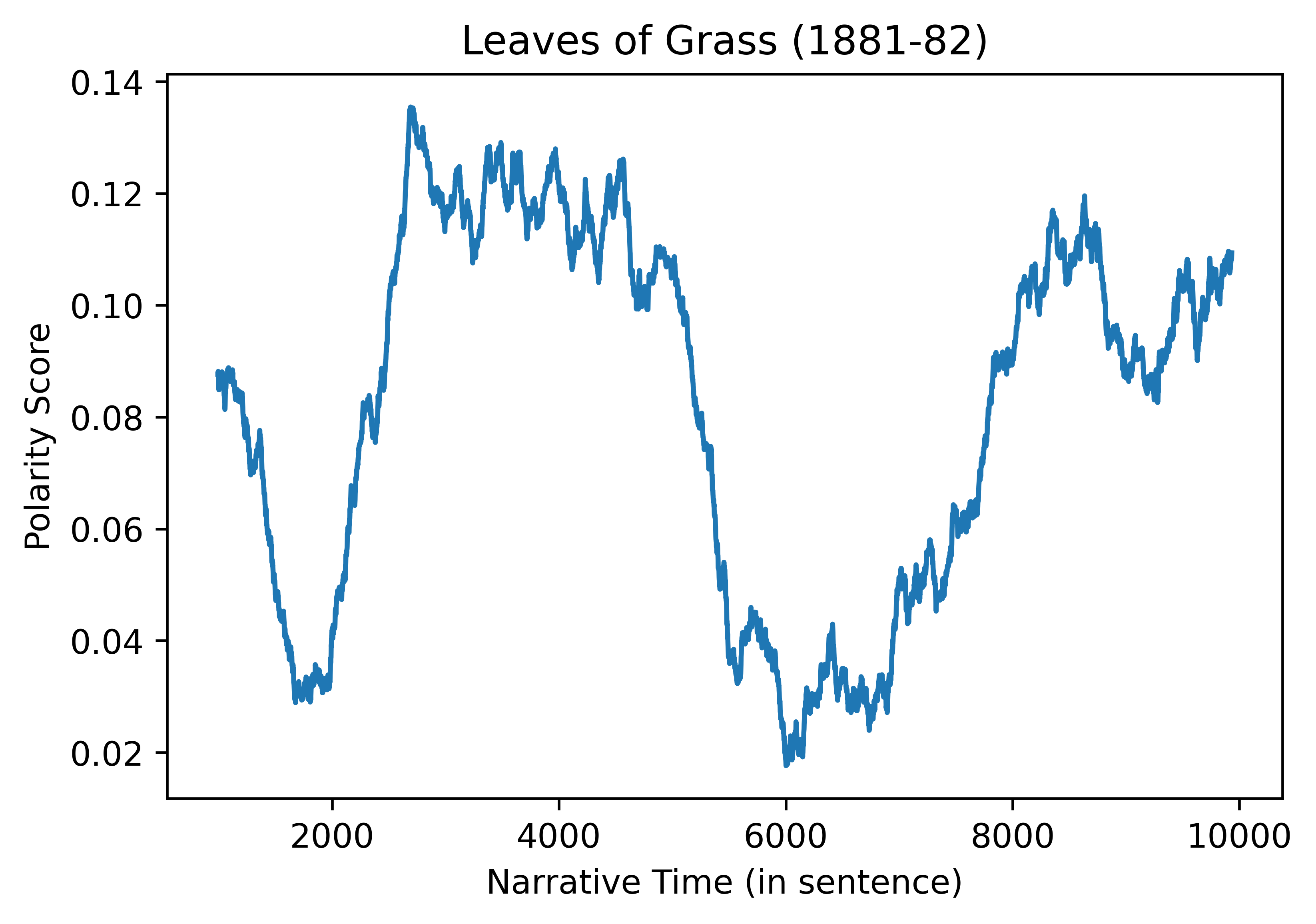}
\end{subfigure}
\begin{subfigure}{0.45\textwidth}
\includegraphics[width=\textwidth]{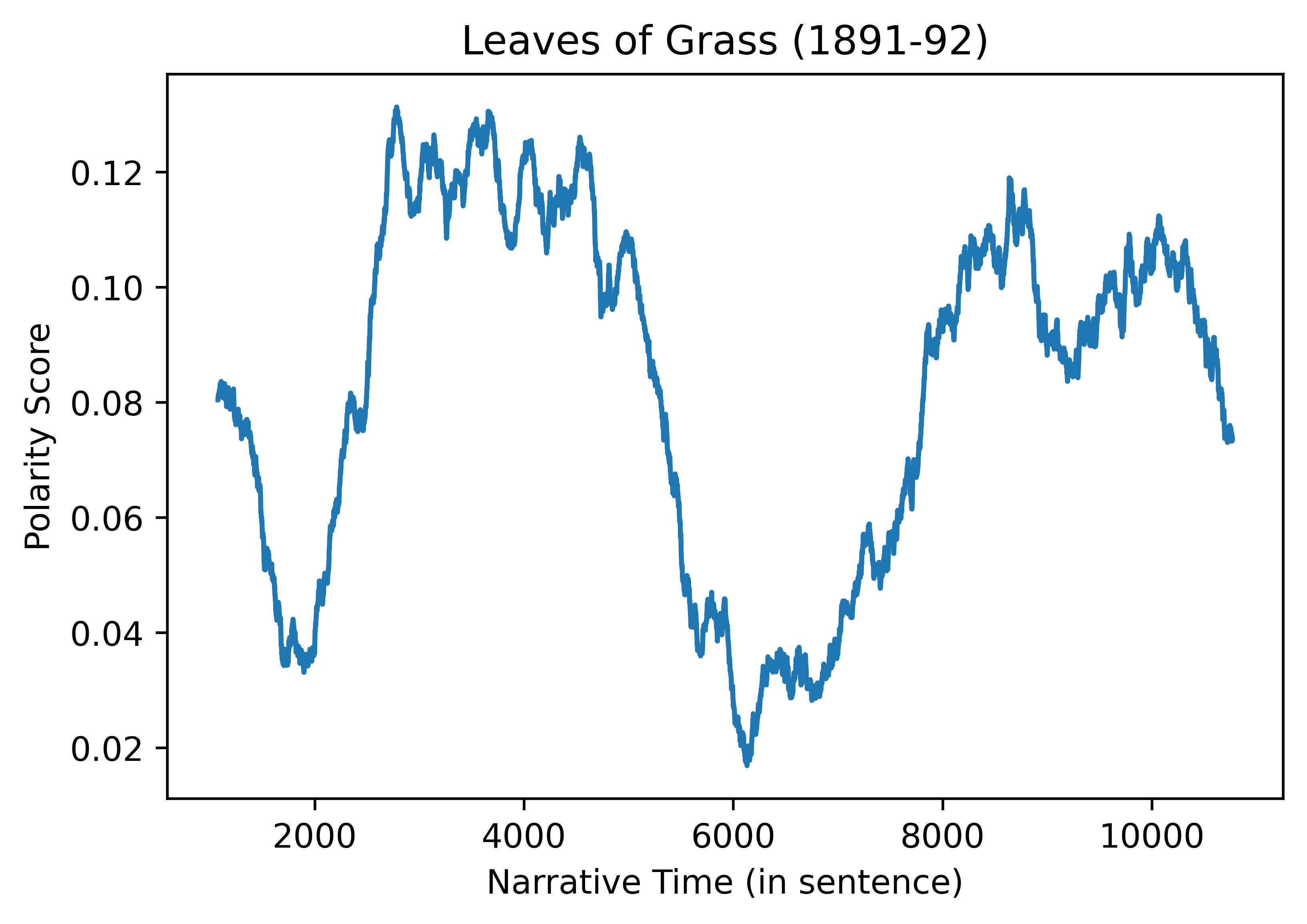}
\end{subfigure}
\caption{Distribution of sentiments in \textit{Leaves of Grass} (rolling mean, window 10\%)}
\label{fig::narrative}
\end{figure}

It is also worthwhile to look at the narrative time of Whitman's sentiments in each edition in Figure \ref{fig::narrative}. Note that we apply a rolling window of 10\% of each text, which means each point represents the rolling mean of polarity scores of 10\% all sentences surrounding it.

\par
It seems that there is a common drop in sentiment after the starting celebratory song of Whitman himself in each edition. We also observe similarity in shape among the 1856, 1860-61, 1867, and 1871-72 edition, which could possibly result from similar featured words in these editions, as shown in Figure \ref{fig::PCA}. The 1881-82 edition is the same with the 1891-92 edition, except that the 1891-92 edition has a downward tail in the end, which derives from the 2 ``Annexes'' added by Whitman.

\clearpage
\section{Conclusion}
From the analysis above, Whitman's writing style changes in editions of \textit{Leaves of Grass}. Some elements of style changed consistently throughout his career, regardless of the Civil War. One example is his increasing favor of French terms, which could possibly be derived from his trip to New Orleans in 1948, long before he published his first edition of \textit{Leaves of Grass}. Similarly, his use of parentheses increased drastically (for about 60 times) in his revisions of the book. Another consistent trend is the increasing use of Latinate words, leading to the gradual increase in lexical complexity. The ever expanding stanza size also contributes to his complicated style in late years.

\par
The Civil War did transform Whitman's poetic style in certain ways. His sentiments in the first postwar edition (the 1867 edition) plunged by around 12 percent as he included postwar trauma in his project of reconstructing \textit{Leaves of Grass}. In fact, all postwar editions are much more pessimistic compared to editions before the war. The cruel war also made Whitman question the meaning of ``self''. Whitman gradually faded away in postwar editions of \textit{Leaves of Grass} as the frequency of self-identity words dropped and more and more ``I''s are replaced by the homophonic ``aye''s.

\par
PCA reveals the landscape of Whitman's stylistic changes. His style changed significantly when he published the first revision of his 1855 edition, as well as when he published the ``definitive'' 1881 edition. The first revision was important for him since the first edition of \textit{Leaves of Grass} did not sell well, and that probably forced him to change his style in the 1856 edition. The 1881 edition was Whitman's unification of his poems, which had a unique style compared to all previous editions, as shown in Figure \ref{fig::PCA}.

\newpage
\theendnotes
\end{document}